\pdfoutput=1

\DeclareUnicodeCharacter{2212}{-}

\documentclass{article}
\usepackage[utf8]{inputenc} 
\PassOptionsToPackage{numbers, sort&compress}{natbib}

    \usepackage[preprint]{neurips_2021}

\usepackage[T1]{fontenc}    
\usepackage{hyperref}       
\usepackage{url}            
\usepackage{booktabs}       
\usepackage{amsfonts}       
\usepackage{nicefrac}       
\usepackage{microtype}      
\usepackage{xcolor}         
\usepackage{bbding}
\usepackage{enumitem}
\usepackage{hyperref}
\usepackage{url}
\usepackage{bm}
\usepackage{graphicx}
\usepackage{booktabs}
\usepackage{multicol}
\usepackage{multirow}
\usepackage{caption}
\usepackage{makecell}
\usepackage{amsmath} 
\usepackage{float}
\usepackage{extarrows}
\usepackage{amssymb}
\usepackage{pifont}
\usepackage{color}
\definecolor{gray}{RGB}{200,200,200}
\definecolor{green}{RGB}{0,200,0}

\title{ CoSP: Co-supervised pretraining of pocket and ligand}
\author{%
  Zhangyang Gao \\
  \And
  Cheng Tan \\
  \And
  Lirong Wu \\
  \And
  Stan Z. Li \footnote{Corresponding author}\\
}

\begin{document}
\maketitle

\begin{abstract}
  Can we inject the pocket-ligand interaction knowledge into the pre-trained model and jointly learn their chemical space? Pretraining molecules and proteins has attracted considerable attention in recent years, while most of these approaches focus on learning one of the chemical spaces and lack the injection of biological knowledge. We propose a co-supervised pretraining (CoSP) framework to simultaneously learn 3D pocket and ligand representations. We use a gated geometric message passing layer to model both 3D pockets and ligands, where each node's chemical features, geometric position and orientation are considered. To learn biological meaningful embeddings, we inject the pocket-ligand interaction knowledge into the pretraining model via contrastive loss. Considering the specificity of molecules, we further propose a chemical similarity-enhanced negative sampling strategy to improve the contrastive learning performance. Through extensive experiments, we conclude that CoSP can achieve competitive results in pocket matching, molecule property predictions,  and virtual screening.
\end{abstract}

\section{Introduction}
Is there a pretrained model that explores the chemical space of pockets and ligands while considering their interactions? Recently, many deep learning methods have been proposed to understand the chemical space of protein pockets or drug molecules (or called ligands) and facilitate drug design in many aspects, e.g., finding hits for a novel target \citep{willmann2012impairment}, repurposing ancient drugs for new targets \citep{weber2004unexpected, kinnings2009drug, yang2015computational}, and searching for similar pockets and molecules \citep{simonovsky2020deeplytough, nguyen2021graphdta}. While deep learning models have shown promising potential in learning separate pocket space or molecular space for specific tasks \citep{stark20213d, liupre, fang2021knowledge, zhang2020motif, gao2022alphadesign}, jointly pretrain pockets and ligands considering their interactions, rather than learning the pocket or molecular spaces individually, remains to be explored.






We propose \textbf{co-s}upervised \textbf{p}retraining (CoSP) framework for understanding the joint chemical space of pockets and ligands. Taking the ligand as an example, contrastive self-supervised pretraining \citep{wang2022molecular, sun2021mocl, fang2021knowledge} has attracted considerable attention and yielded significant achievements in recent years. By identifying expertly defined positive and negative ligand pairs, the model can learn the underlying knowledge to facilitate downstream tasks. However, most self-supervised methods capture data dependencies in the "self" data domain while ignoring additional information from other complementary fields, such as bindable pockets. To fill this knowledge gap, we introduce cross-domain dependencies between pockets and ligands to optimize the representations. The basic assumption is that functionally similar molecules prefer to bind to similar pockets and vice versa.








We propose \textbf{g}ated \textbf{g}eometric \textbf{m}assage \textbf{p}assing (GGMP) layer to extract expressive bio-representations for 3D pockets and ligands. All bio-objects are treated as 3D graphs in that each node contains invariant chemical features (atomic number, etc.) and equivalent geometric features (position and direction). For each bio-object, we analyze the pairwise energy function \citep{guan2021energy}, which considers both chemical features and geometric features via the gated operation. By minimizing the energy function, we derive the updating rules of position and direction vectors. Finally, we combine these rules with classical message passing, resulting in GGMP.





We introduce ChemInfoNCE loss to reduce the negative sampling bias. When applying contrastive learning, the false negative pairs that are actually positive will lead to performance degradation, called negative sampling bias.  Chuang et al. \citep{chuang2020debiased} assume that the label distribution of the classification task is uniform and propose Debiased InfoNCE to alleviate this problem. Considering the specificity of the molecules and extending the situation to continuous properties prediction (regression task), we propose the chemical similarity-enhanced negative ligand sampling strategy. Interestingly, the change in sampling is equivalent to the change in the sample weights. Form the view of loss functions, we provide a systematic understanding and propose ChemInfoNCE.

We evaluates our model on several downstream tasks, from pocket matching, molecule property prediction to virtual screening. Numerous experiments show that our approach can achieve competitive results on these tasks, which may provide a good start for future AI drug discovery.

\section{Related work}

\paragraph{ Motivation } Protein and molecule show their biological functions by binding with each other \citep{chaffey2003alberts}, thus exploring the protein-ligand complex help to improve the understanding of both proteins, molecules, and their interactions. To improve the generalization ability and reduce the complexity, we further consider local patterns about the protein pocket $x$ and its ligand $\hat{x}$. Taking $(x,\hat{x})$ as the positive pair, while $(x, \hat{x}^-)$ as the negative pair, where $\hat{x}^-$ could not bind to $x$, we aims to pretrain pocket model $f: x \mapsto \boldsymbol{h}$ and a ligand model $\hat{f}: \hat{x} \mapsto \hat{\boldsymbol{h}}$, such that the mutual information between $\boldsymbol{h}$ and $\hat{\boldsymbol{h}}$ will be maximized.

\paragraph{Equivalent 3D GNN} Extensive works have shown that 3D structural conformation can improve the quality of bio-representations with the help of equivalent massage passing layer \citep{cohen2016group, thomas2018tensor, fuchs2020se, satorras2021n, brandstetter2021geometric, batzner20223}. Inspired by the energy analysis \citep{ganea2021independent, guan2021energy}, we propose a new \textbf{g}ated \textbf{g}eometric \textbf{m}assage \textbf{p}assing (GGMP) layer that consider not only the node position but also its direction, where the latter could indicate the cavity position of the pocket and describe the 3D rotational state.



\paragraph{InfoNCE} The original InfoNCE is proposed by \citep{oord2018representation} to contrast semantically similar (positive) and dissimilar (negative) pairs of data points, such that the representations of similar pairs $(x, \hat{x})$ to be close, and those of dissimilar pairs $(x, \hat{x}^-)$ to be more orthogonal. By default, the negative pairs are uniformly sampled from the data distribution. Therefore, false negative pairs will lead to significant performance drop. To address this issue, DebaisedInfoNCE\citep{chuang2020debiased} is proposed, which assumes that the label distribution of the classification task is uniform. Although DebaisedInfoNCE has achieved good results on image classification, it is not suitable for direct transfer to regression tasks, and the uniform distribution assumption may be too strict. As to the bio-objects, we discard the above assumption, extend the situation to continuous attribute prediction, use fingerprint similarity to measure the probability of negative ligands, and propose ChemInfoNCE.

\paragraph{Self Bio-Pretraining} Many pre-training methods have been proposed  for a single protein or ligand single domain, which can be classified as sequence-based, graph-based or structure-based. We summarize the \textbf{protein pretraining} models in Table.\ref{tab:pretrain}. As for sequential models, CPCPort \citep{lu2020self} maximizes the mutual information between predicted residues and context. Profile Prediction \citep{sturmfels2020profile} suggests predicting MSA profile as a new pretraining task. OntoProtein \citep{zhang2022ontoprotein} integrates GO (Gene Ontology) knowledge graphs into protein pre-training. While most of the sequence models rely on the transformer architecture, CARP \citep{yang2022convolutions} finds that CNNs can achieve competitive results with much fewer parameters and runtime costs. Recently, GearNet \citep{zhang2022protein} explores the potential of 3D structural pre-training from the perspective of masked prediction and contrastive learning. We also summarize the \textbf{molecule pretraining} models in Table.\ref{tab:pretrain}. As for sequential models, FragNet \citep{shrivastava2021fragnet} combines masked language model and multi-view contrastive learning to maximize the inner mutual information of the same SMILEs and the agreement across augmented SMILEs. Beyond SMILEs, more approaches \citep{rong2020self, li2021effective, sun2021mocl, fang2021knowledge, zhang2020motif, zhang2021motif, wang2022molecular} tend to choose graph representation that can better model structural information. For example, Grover \citep{rong2020self} integrates message passing and transformer architectures and pre-trains a super-large GNN using 10 million molecules. MICRO-Graph \citep{zhang2020motif} and MGSSL \citep{zhang2021motif} use motifs for contrastive learning. Considering the domain knowledge, MoCL \citep{sun2021mocl} uses substructure substitution as a new data augmentation operation and predicts pairwise fingerprint similarities. Although these pre-training methods show promising results, they do not consider the 3D molecular conformations. To fill this gap, GraphMVP \citep{liupre} and 3DInfomax \citep{stark20213d} explore to maximize the mutual information between 3D and 2D views of the same molecule and achieve further performance improvements. Besides, GEM \citep{fang2022geometry} proposes a geometry-enhanced graph neural network and pretrains it via geometric tasks. For the pretraining of individual proteins or molecules, these methods demonstrate promising potential on various downstream tasks but ignore their interactions.

\begin{table}[h]
  \centering
  \resizebox{1.0 \columnwidth}{!}{
  \begin{tabular}{llll|llll}
  \toprule
  \multicolumn{4}{c|}{Protein} & \multicolumn{4}{c}{Molecule}  \\
  Method              & Data & Code & Year   & Method       & Data & Code & Year\\
  \midrule
  CPCProt  \citep{lu2020self}           & sequence    &   \href{https://github.com/amyxlu/CPCProt.git}{PyTorch}  & 2020  & FragNet \citep{shrivastava2021fragnet} & SMILEs & -- & 2021\\

  Profile Prediction  \citep{sturmfels2020profile}    &  sequence   &   --  & 2020 & MoCL \citep{sun2021mocl} & graph & \href{https://github.com/illidanlab/MoCL-DK.git}{PyTorch} & 2021\\

  ONTOPROTEIN  \citep{zhang2022ontoprotein}   & sequence   &   \href{https://github.com/zjunlp/OntoProtein}{PyTorch}   & 2022 & MPG \citep{li2021effective} & graph & \href{https://github.com/pyli0628/MPG.git}{PyTorch} & 2021\\

  CARP                   & sequence   &  --  & 2022 & Grover \citep{rong2020self}         &  graph    &  \href{https://github.com/tencent-ailab/grover.git}{PyTorch}  & 2020 \\
  GearNet                   &   3D   &   --   & 2022 & MICRO-Graph \citep{zhang2020motif} & graph & -- & 2020\\
   & & & & CKGNN \citep{fang2021knowledge} & graph & -- & 2021\\
   & & & & MGSSL \citep{zhang2021motif} & graph & \href{https://github.com/zaixizhang/MGSSL}{PyTorch} & 2021\\
   & & & & MolCLR \citep{wang2022molecular} & graph & \href{https://github.com/yuyangw/MolCLR}{PyTorch} & 2022\\
   & & & & 3DInfomax \citep{stark20213d} & graph+3D & \href{https://github.com/HannesStark/3DInfomax.git}{PyTorch} & 2021\\
   & & & & GraphMVP \citep{liupre} & graph+3D & -- &2022 \\
   & & & & GEM \citep{fang2022geometry} & graph+3D & \href{https://github.com/PaddlePaddle/PaddleHelix/tree/dev/apps/pretrained_compound/ChemRL/GEM}{Paddle} & 2022\\
  \bottomrule
  \end{tabular}}
  \caption{Protein and molecule pretraining methods}
  \label{tab:pretrain}
\end{table}

\paragraph{Cross Bio-Pretraining} In parallel with our study, Uni-Mol \citep{zhou2022uni}, probably the first pretrained model that can handle both protein pockets and molecules, released the preprinted version. Compared with them, our approach differs in pretrained data, pretraining strategy, model architecture, and downstream tasks.

\section{Methodology}
\subsection{Co-supervised pretraining framework}
\begin{figure}[h]
  \centering
  \includegraphics[width=5.5in]{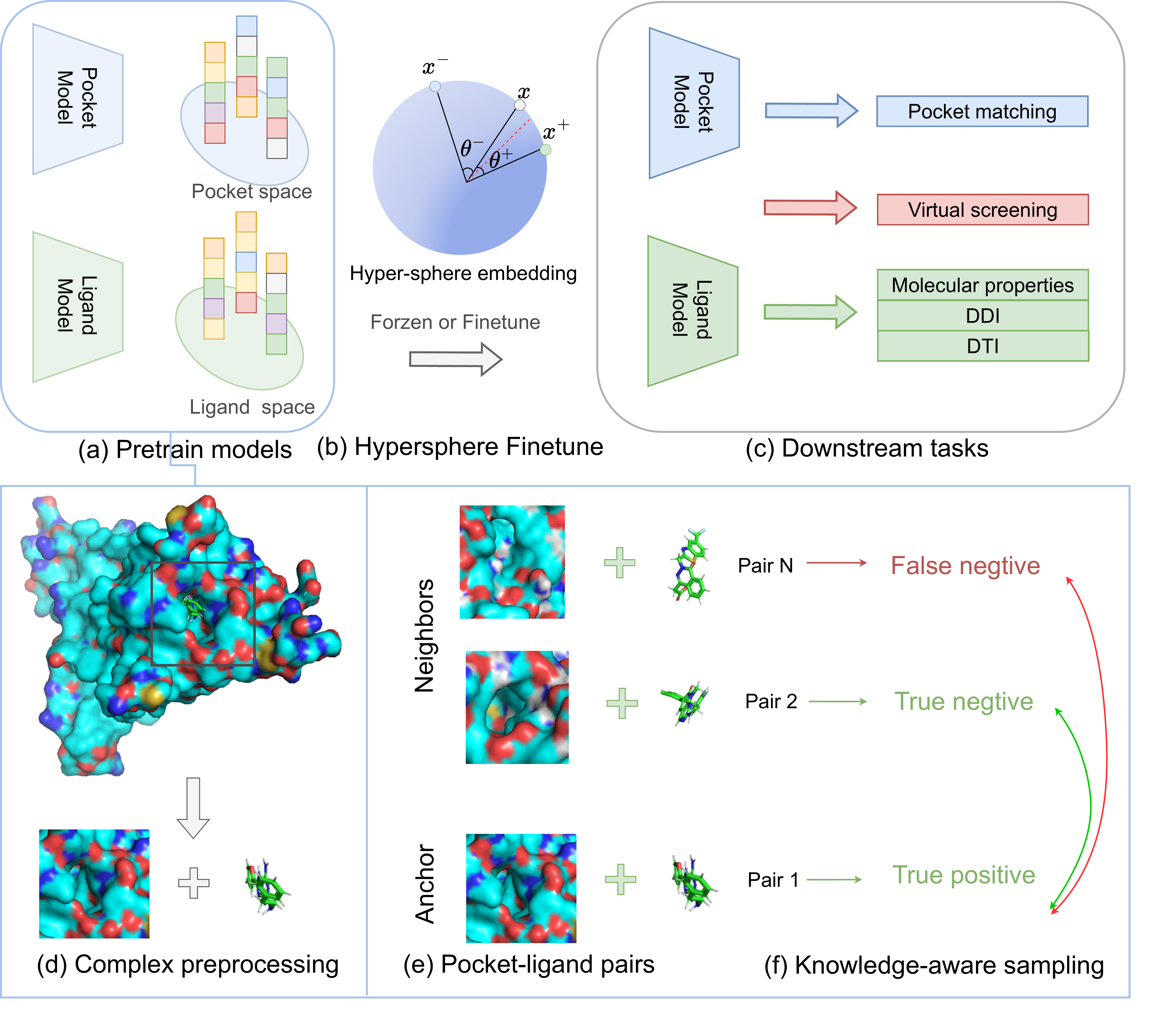}
  \caption{ Overview of CoSP. We contrast bound pocket-ligand pairs with unbound ones to learn the interaction-aware chemical embeddings.}
  \label{fig:global_structure}
\end{figure}

We propose the \textbf{co-s}upervised \textbf{p}retraining (CoSP) framework, as shown in Figure.\ref{fig:global_structure}, to explore the joint chemical space of protein pockets and ligands, where the methodological innovations include:

\begin{itemize}[leftmargin=5mm]
    \item We propose the gated geometric message passing layer to model 3D pockets and ligands.
    \item We propose a co-supervised pretraining framework to learn pocket and ligand representations.
    \item We propose ChemInfoNCE with improved negative sampling guided by chemical knowledge.
    \item We evaluate the pretrained model on both pocket matching, molecule property prediction, and virtual-screening tasks.
\end{itemize}




\subsection{Geometric Representation}
We introduce the unified data representation and neural network for modeling 3D pockets and ligands. All structures used for pre-training are collected from the BioLip dataset. In downstream tasks where ligand conformations are not provided, we generate 3D conformations using MMFF (if successful) or their 2D conformations (if failed).

\paragraph{Pocket and Ligand Graph}
We represent bio-object as graph $\mathcal{G}(X, \mathcal{V}, \mathcal{E})$ , consisting of coordinate matrix $X \in \mathbb{R}^{n,3}$, node features $\mathcal{V} \in \mathbb{R}^{n, d_f}$ , and edge features $\mathcal{E} \in \mathbb{R}^{n, d_e}$. For pockets, the graph nodes include amino acids within 10$\mathring{A}$ to the ligand, $X$ contrains the position of $C_{\alpha}$ of residues, on which we construct $\mathcal{E}$ via k-nn algorithm. For molecules, the graph nodes include all ligand atoms except Hs, $X$ contrains the atom positions, and we use the molecular bonds as $\mathcal{E}$.

\paragraph{Gated Geometric Massage Passing} From layer $t$ to $t+1$, we use the gated geometric massage passing (GGMP) layer to update 3D graph representations, i.e., $[\boldsymbol{v}_i^{t+1}, \boldsymbol{x}_{i}^{t+1} , \boldsymbol{n}_{i}^{t+1}] = \mathrm{GGMP}(\boldsymbol{v}_i^{t},  \boldsymbol{x}_{i}^{t}, \boldsymbol{n}_{i}^{t})$, where $\boldsymbol{n}_{i}$ is the direction vector that point to the neighborhood center of node $i$. For protein pockets, $\boldsymbol{n}_{i}$ could indicate the position of protein caves. As to the 3D conformations, we minimize the pairwise energy function $E(X, F, \mathcal{E})$:

\begin{align}
  E(X, F, \mathcal{E}) = \sum_{(i,j) \in \mathcal{E}} u(\boldsymbol{v}_i, \boldsymbol{v}_j, \boldsymbol{e}_{ij}) g(\langle \boldsymbol{n}_i, \boldsymbol{n}_j \rangle, d_{ij}^2)
\end{align}

where $d_{ij}^2 = ||\boldsymbol{x}_i-\boldsymbol{x}_j||^2$, both chemical energy $u(\cdot)$ and geometric energy $g(\cdot)$ are considered. By calculating the gradients of $\boldsymbol{x}_i$ and $\boldsymbol{n}_i$, we get their updating rules: 


\begin{align}
  -\frac{\partial E(X, F, \mathcal{E})}{\partial \boldsymbol{x}_i} 
  & = -\sum_{j \in \mathcal{N}_i} 2 u_{ij} \frac{\partial g_{ij} }{\partial d_{ij}^2} (\boldsymbol{x}_i-\boldsymbol{x}_j) 
   \approx \sum_{j \in \mathcal{N}_i} u(\boldsymbol{v}_i, \boldsymbol{v}_j, \boldsymbol{e}_{ij}) \phi_x(d_{ij}^2, \langle \boldsymbol{n}_{i}^{t}, \boldsymbol{n}_{j}^{t} \rangle) (\boldsymbol{x}_{i}^{t}-\boldsymbol{x}_{j}^{t})\\
  -\frac{\partial E(X, F, \mathcal{E})}{\partial \boldsymbol{n}_i} 
  & = -\sum_{j \in \mathcal{N}_i} u_{ij} \frac{\partial g_{ij} }{\partial \langle \boldsymbol{n}_i, \boldsymbol{n}_j \rangle} \boldsymbol{n}_j 
   \approx \sum_{j \in \mathcal{N}_i} u(\boldsymbol{v}_i, \boldsymbol{v}_j, \boldsymbol{e}_{ij}) \phi_n(d_{ij}^2, \langle \boldsymbol{n}_{i}^{t}, \boldsymbol{n}_{j}^{t} \rangle) \boldsymbol{n}_{j}^{t} \\
\end{align}

Note that $\phi_x$ and $\phi_n$ are the approximation of $\frac{\partial g_{ij} }{\partial d_{ij}^2}$ and $\frac{\partial g_{ij} }{\partial \langle \boldsymbol{n}_i, \boldsymbol{n}_j \rangle}$. Combining graph message passing, we propose the GGMP:

\begin{align}
  & \boldsymbol{m}_{ij} = \phi_{m}(\boldsymbol{v}_i^t, \boldsymbol{v}_j^t, e_{ij} )\\
  & \boldsymbol{g}_{ij} = \phi_{g}(d_{ij}^2, \langle \boldsymbol{n}_{i}^{t}, \boldsymbol{n}_{j}^{t} \rangle )\\
  & \boldsymbol{h}_i^{t+1} = \phi_{h}(\boldsymbol{h}_i^t, \sum_{j \in \mathcal{N}_i} \boldsymbol{m}_{ij} \boldsymbol{g}_{ij})\\ 
  & \boldsymbol{x}_{i}^{t+1} = \boldsymbol{x}_{i}^{t} + \lambda \sum_{j\in \mathcal{N}_{i}} u(\boldsymbol{m}_{ij}) \phi_x(\boldsymbol{g}_{ij}) (\boldsymbol{x}_{i}^{t}-\boldsymbol{x}_{j}^{t}) \\
  & \boldsymbol{n}_{i}^{t+1} = \boldsymbol{n}_{j}^{t} + \lambda \sum_{j\in \mathcal{N}_{i}} u(\boldsymbol{m}_{ij}) \phi_n(\boldsymbol{g}_{ij}) \boldsymbol{n}_{j}^{t} \\
\end{align}

where $\phi_*$ and $u$ are approximated by neural networks. As suggested in \citep{ganea2021independent,guan2021energy}, the neural layer could learn expressive representations by considering both geometric constraints and chemical features.




\subsection{Contrastive loss}


In contrastive learning, the biased negative sampling could impair the performance, in which case false negative pairs are sampled for training. Previous methods \citep{chuang2020debiased, robinson2020contrastive} address this problem by simply considering the class distribution of false-negative samples as uniform under the classification setting. We propose chemical knowledge-based sampling to better address this issue, where fingerprint similarity is used to measure the probability of negative ligands. Interestingly, the change in sampling distribution is actually equivalent to the design of a weighted loss, and we provide a comprehensive understanding from the perspective of contrastive loss.


\paragraph{Uni-contrastive loss} Given the pocket $x \sim p$, we draw positive ligands $\hat{x}^+$ from the distribution $\hat{p}_x^+$ of bindable molecules and negative ligands $\{\hat{x}_i^-\}_{i=1}^N $ from the distribution $\hat{q}$ of non-bindable ones. By default, the positive ligands are determined by the pocket-ligand complexes, while negative are uniformly sampled from the ligand sets. We use pocket model $f$ and ligand model $\hat{f}$ to pockets and ligands into the latent space, resulting in $\boldsymbol{h}$, $\hat{\boldsymbol{h}}^+$ and $\{ \hat{\boldsymbol{h}}_i^- \}_{i=1}^N$, and maximize the positive similarity $s^+=\text{sim}(\boldsymbol{h}, \hat{\boldsymbol{h})}$ against the negative similarities $s_{i}^- = \text{sim}(\boldsymbol{h} ,\hat{\boldsymbol{h}}_i^-), i=1,2,\cdots$, resulting in:




\begin{align}
  L_{\text{Uni}} &= \mathbb{E}_{x \sim p, \hat{x}^+ \sim \hat{p}_x^+, \atop \{\hat{x}_i^-\}_{i=1}^N \sim \hat{q}}
  \left[ \log{(1 + \frac{Q}{N} \sum_{i=1}^N s_i^- / s^+)} \right]\\
\end{align}


For each data sample $x$, the gradients contributed to $s^+$ and $s_i^-$ are:

\begin{align}
  \frac{\partial{L}}{\partial s^+} &= \frac{1}{1+\sum_{i=1}^N s_i^- / s^+} \sum_{i=1}^N {\frac{\partial s_i^-/s^+}{\partial s^+}}\\
  \frac{\partial{L}}{\partial s_i^-} &= \frac{1}{1+\sum_{i=1}^N s_i^- / s^+}  \frac{\partial s_i^-/s^+}{\partial s_i^-}
\end{align}

One can verify that InfoNCE is the special case of $L_{\text{Uni}}$ by setting $s^+ = e^{\gamma \boldsymbol{h}^T \boldsymbol{h}^+}$ and $s_i^- = e^{\gamma \boldsymbol{h}^T \boldsymbol{h}_i^-}$.

\paragraph{DebiasedInfoNCE} \citep{chuang2020debiased} points out sampling from the data distribution $\hat{q}$ will harm the model performance because the positive samples may be mistaken for negative ones. Denote $h(\cdot)$ as the labeling function, they suggest to draw negative samples from $\hat{q}_x^-(\hat{x}^-) = p(\hat{x}^-|h(\hat{x}^-) \neq h(x))$, which can be viewed as the real distribution of negative ligands. To study the event of $\{ h(\hat{x}^-) \neq h(x) \}$, they further consider the joint distribution $p(\hat{x},c)=p(\hat{x}|c)p(c)$ over data $x$ and label $c$. By assuming the class probability $p(c)=\tau^+$ as uniform, and let $\tau^-=1-\tau^+$ as the probability of observing any different class, they decompose $\hat{q}$ as $\tau^- \hat{q}_x^-(x^-) + \tau^+ \hat{q}_x^+(x^-)$ and propose

\begin{align}
  L_{\text{Debiased}} &= \mathbb{E}_{x \sim p, \hat{x}^+ \sim \hat{p}_x^+, \atop \{\hat{x}_i^-\}_{i=1}^N \sim \hat{q}_x^-}
  \left[ \log{(1 + \frac{Q}{N} \sum_{i=1}^N s_i^- / s^+)} \right]\\
\end{align}

By choosing $s^+=e^{\boldsymbol{h}^T \hat{\boldsymbol{h}}^+}$, $s_i^-=e^{\boldsymbol{h}^T \hat{\boldsymbol{h}}_i^-}$, and replace $\hat{q}_x^-(x^-) = (\hat{q} - \tau^+ \hat{q}_x^+)/\tau^-$, with mild assumptions, the approximated debaised InfoNCE can be written as:

\begin{align}
  L_{\text{Debiased}} & \approx \mathbb{E}_{x \sim p, x^+ \sim p_x^+} \left[ \log{ (1+ \frac{1}{\tau^-}   \sum_{i=1}^{N} (e^{\boldsymbol{h}^T \boldsymbol{h}_i^- - \boldsymbol{h}^T \boldsymbol{h}^+} - \tau^+)  ) }\right]
\end{align}

\paragraph{ChemInfoNCE} For the case of classification with discrete labels, the assumption of uniform class probabilities may be too strong, especially for the unbalanced dataset. When it comes to regression, molecules have continuous chemical properties and the event $\{ h(\hat{x}) \neq h(\hat{x}^-) \}$ can not describe the validity of negative data. Considering these challenges, we introduce a new event $\{ \text{score}(\hat{x},\hat{x}^-)<\tau \}$ to measure the validity of negative samples, where $\text{score}(\cdot)$ is the function of chemical similarity. The underlying hypothesis is that molecules with lower chemical semantics are more likely to be negative samples. Thus, we define the true negative probability as:

\begin{align}
  q_x^-(\hat{x}^-) := q(\hat{x}^-| \text{sim}(x, \hat{x}^-) < \tau) \propto \max(1-\text{sim}(x,\hat{x}^-)-\tau,0 ) \cdot p(\hat{x}^-)
\end{align}

Bt denoting $w_i = \max(1-\text{sim}(x,\hat{x}^-)-\tau,0 )$, similar to DebaisedInfoNCE, the final ChemInfoNCE can be simplfied as:








\begin{align}
  L_{\text{Chem}} & \approx \mathbb{E}_{x \sim p, x^+ \sim p_x^+} \left[ \log{ (1+   \sum_{i=1}^{N} ( \rho_i e^{\boldsymbol{h}^T \boldsymbol{h}_i^- - \boldsymbol{h}^T \boldsymbol{h}^+} )  ) }\right]
\end{align}
where $\rho_i = \frac{w_i}{\sum_{i=1}^N w_i}$. We provide a detailed derivation in the appendix.

\section{Experiments}
In this section, we conduct extensive experiments to reveal the effectiveness of the proposed method from three perspectives:
\begin{enumerate}[leftmargin=8mm]
  \item \textbf{Pocket}: How does the pre-trained pocket model perform on the pocket matching tasks?
  \item \textbf{Ligand}: Could the ligand model provide good performance in predicting molecular properties?
  \item \textbf{Pocket-ligand}: Can the joint model improve the performance of finding potentially binding pocket-ligand pairs, i.e., virtual screening?
\end{enumerate}




\subsection{Pretraining setup}
\paragraph{Pretraining Dataset}
We adopt BioLip \citep{yang2012biolip} dataset for co-supervised pretraining, where the original BioLip contains 573,225 entries up to 2022.04.01. Compared to PDBBind \citep{wang2005pdbbind} with 23,496 complexes, BioLip contains more complexes that lack binding affinity, thus could provide a more comprehensive view of binding mode analysis. To focus on the drug-like molecules and their binding pockets, we filtered out other unrelated complexes that contain pipelines, DNA, RNA, single ions, etc. The whole data preprocessing pipeline can be found in the appendix.


\paragraph{Experimental setting} We pretrain stacked GGMPs via ChemInfoNCE loss to minimize the distance of positive pairs, while maximizing the distance of negative pairs. Experiments are conducted on NVIDIA A100s, where the learning rate is 0.01 and the batch size is 100. 


\subsection{Downstream task 1: Pocket matching}

\paragraph{Experimental setup} Could the pre-trained model identify chemically similar pockets from various pockets? We explore the discriminative ability of the pocket model with the pocket matching tasks. To comprehensively understand the potential of the proposed method, we evaluated it on 10 benchmarks recently collected in the ProSPECCTs dataset \citep{ehrt2018benchmark}. For each benchmark, the positive and negative pairs of pockets are defined differently according to the research objectives. We summarize 5 research objectives as \textbf{O1}: Whether the model is robust to the pocket definition? \textbf{O2}: Whether the model is robust to the pocket flexibility? \textbf{O3}: Can the model distinguish between pockets with different properties? \textbf{O4}: Whether the model can distinguish dissimilar proteins binding to identical ligands and cofactors? \textbf{O5}: How about the performance on real applications? The motivation and dataset descriptions of these objectives can be found in Table.~\ref{tab: dataset_pocket}. We report the AUC-ROC scores on all benchmarks.

\begin{table}[H]
  \small
  \begin{tabular}{p{0.5cm} p{3cm} p{7cm} p{1.5cm}}
    \toprule
  Name       & Goal & Description & $\frac{N_{pos}}{N_{neg}}$\\
  \midrule
  D1        & Pocket definition. & D1 contrains structures with identical sequences. Since the same pocket could bind to different ligands, diverse pocket definition will influce the learned representations. 
  & 13430/92846\\
  D1.2   & Impact of ligand diversity. & D1.2 is a reduced set of D1 that exclusively contains structures with identical sequences and similar ligands. & 241/1784\\
  D2         & Conformational flexibility. & D2 uses NMR structures that provide conformational ensembles of protein structures, where the same pockets contain larger and smaller conformational variations. & 7729/100512 \\
  D3         & Physicochemical properties & Based on D1, D3 further adds negative samples by replacing pockets' residues with ones that have different physicochemical properties and similar size. & 13430/67150 \\
  D4         & Physicochemical and shape properties. & Based on D1, D3 further adds negative samples by replacing pockets' residues with ones whose number of carbon and hetero atoms differs by at least three atoms. & 13430/67150\\
  D5 & Impact of binding site features, interaction patterns & Kahraman data set without phosphate binding sites & 920/5480  \\
  D5.2 &Similar to D5. & Kahraman data set & 1,320/8,680 \\
  D6 & Relationships between pockets binding to identical ligands. & Barelier data set & 19/43 \\
  D6.2 & Similar to D6. & Barelier data set including cofactors & 19/43 \\
  D7   & Successful applications      & D7 consists of pocket pairs which were previously characterized as being similar in published literature & 115/56284\\
  \bottomrule 
  \end{tabular}
  \caption{Pocket matching benchmarks and objectives.}
  \label{tab: dataset_pocket}
  \end{table}

\paragraph{Baselines} We compare the proposed approach with both classical and deeplearning baselines. The classical methods can be divided into profile-based, graph-based and grid-based ones. The profile-based methods encode topological, physicochemical and statistical properties in a unified way for comparing various pockets, e.g., SiteAlign \citep{schalon2008simple}, RAPMAD \citep{krotzky2015large}, TIFP \citep{desaphy2013encoding}, FuzCav \citep{weill2010alignment}, PocketMatch\citep{yeturu2008pocketmatch}, SMAP\citep{xie2008detecting}, TM-align\citep{zhang2005tm}, KRIPO\citep{wood2012pharmacophore} and Grim \citep{desaphy2013encoding}. The graph-based methods  adopt isomorphism detection algorithm to find the common motifs between pockets, e.g., Cavbase \citep{schmitt2002new}, IsoMIF\citep{chartier2015detection}, ProBiS\citep{konc2010probis}. Grid-based methods represent protein pockets by regularly spaced pharmacophoric grid points, e.g.,VolSite/Shaper \citep{desaphy2012comparison}. Another tools such as SiteEngines\citep{shulman2005siteengines} and SiteHopper\citep{batista2014sitehopper} are also included. As to deeplearning approaches, we compare with the recent SOTA algorithm--DeeplyTough\citep{simonovsky2020deeplytough}.




\paragraph{Results and analyses} We present the pocket matching results in Table.~\ref{tab: pocket_match}, where the pretrained model achieves competitive results in most cases. Specifically, CoSP is robust to pocket definition (\textbf{O1}) and achieves the highest AUC scores in D1 and D1.2. The robustness also remains when considering conformational variability (\textbf{O2}), where CoSP achieves 0.99 AUC score in D2. It should be noted that robustness to homogeneous pockets does not mean that the model has poor discrimination; on the contrary, the model could identify pockets with different physicochemical and shape properties (\textbf{O3}) in D3 and D4. Compared with previous deep learning methods (DeeplyTough), CoSP provides better performance in distinguishing different pockets bound to the same ligands and cofactors (\textbf{Q4}), refer to the results of D5, D5.2, D6 and D6.2. Last but not least, CoSP showed good potential for practical applications (\textbf{O5}) with 0.81 AUC score.



\begin{table}[h]
  \small
  \resizebox{1.0 \columnwidth}{!}{
  \begin{tabular}{lcccccccccc}
  \toprule
                       & D1   & D1.2 & D2   & D3   & D4   & D5   & D5.2 & D6   & D6.2 & D7   \\ \midrule
  Cavbase              & 0.98 & 0.91 & 0.87 & 0.65 & 0.64 & 0.60 & 0.57 & 0.55 & 0.55 & 0.82 \\
  FuzCav               & 0.94 & 0.99 & 0.99 & 0.69 & 0.58 & 0.55 & 0.54 & 0.67 & 0.73 & 0.77 \\
  FuzCav (PDB)         & 0.94 & 0.99 & 0.98 & 0.69 & 0.58 & 0.56 & 0.54 & 0.65 & 0.72 & 0.77 \\
  grim                 & 0.69 & 0.97 & 0.92 & 0.55 & 0.56 & 0.69 & 0.61 & 0.45 & 0.65 & 0.70 \\
  grim (PDB            & 0.62 & 0.83 & 0.85 & 0.57 & 0.56 & 0.61 & 0.58 & 0.45 & 0.50 & 0.64 \\
  IsoMIF               & 0.77 & 0.97 & 0.70 & 0.59 & 0.59 & 0.75 & 0.81 & 0.62 & 0.62 & 0.87 \\
  KRIPO                & 0.91 & 1.00 & 0.96 & 0.60 & 0.61 & 0.76 & 0.77 & 0.73 & 0.74 & 0.85 \\
  PocketMatch          & 0.82 & 0.98 & 0.96 & 0.59 & 0.57 & 0.66 & 0.60 & 0.51 & 0.51 & 0.82 \\
  ProBiS               & 1.00 & 1.00 & 1.00 & 0.47 & 0.46 & 0.54 & 0.55 & 0.50 & 0.50 & 0.85 \\
  RAPMAD               & 0.85 & 0.83 & 0.82 & 0.61 & 0.63 & 0.55 & 0.52 & 0.60 & 0.60 & 0.74 \\
  shaper               & 0.96 & 0.93 & 0.93 & 0.71 & 0.76 & 0.65 & 0.65 & 0.54 & 0.65 & 0.75 \\
  shaper (PDB)         & 0.96 & 0.93 & 0.93 & 0.71 & 0.76 & 0.66 & 0.64 & 0.54 & 0.65 & 0.75 \\
  VolSite/shaper       & 0.93 & 0.99 & 0.78 & 0.68 & 0.76 & 0.56 & 0.58 & 0.71 & 0.76 & 0.77 \\
  VolSite/shaper (PDB) & 0.94 & 1.00 & 0.76 & 0.68 & 0.76 & 0.57 & 0.56 & 0.50 & 0.57 & 0.72 \\
  SiteAlign            & 0.97 & 1.00 & 1.00 & 0.85 & 0.80 & 0.59 & 0.57 & 0.44 & 0.56 & 0.87 \\
  SiteEngine           & 0.96 & 1.00 & 1.00 & 0.82 & 0.79 & 0.64 & 0.57 & 0.55 & 0.55 & 0.86 \\
  SiteHopper           & 0.98 & 0.94 & 1.00 & 0.75 & 0.75 & 0.72 & 0.81 & 0.56 & 0.54 & 0.77 \\
  SMAP                 & 1.00 & 1.00 & 1.00 & 0.76 & 0.65 & 0.62 & 0.54 & 0.68 & 0.68 & 0.86 \\
  TIFP                 & 0.66 & 0.90 & 0.91 & 0.66 & 0.66 & 0.71 & 0.63 & 0.55 & 0.60 & 0.71 \\
  TIFP (PDB)           & 0.55 & 0.74 & 0.78 & 0.56 & 0.57 & 0.54 & 0.53 & 0.56 & 0.61 & 0.66 \\
  TM-align             & 1.00 & 1.00 & 1.00 & 0.49 & 0.49 & 0.66 & 0.62 & 0.59 & 0.59 & 0.88 \\
  DeeplyTough          & 0.95 & 0.98 & 0.90 & 0.76 & 0.75 & 0.67 & 0.63 & 0.54 & 0.54 & 0.83 \\ \hline
  CoSP    & 1.00 & 1.00 & 0.99 & 0.79 & 0.81 & 0.62 & 0.63 & 0.61 & 0.62 & 0.81\\ 
  \bottomrule
  \end{tabular}}
  \caption{Pocket matching results.}
  \label{tab: pocket_match}
  \end{table}

\subsection{Downstream task 2: Molecule property prediction}
\paragraph{Experimental setup} Could the model learn expressive features for molecule classification and regression tasks? We evaluate CoSP on 9 benchmarks collected by MoleculeNet\citep{wu2018moleculenet}. Following previous researches, we use scaffold splitting to generate train/validation/test set with a ratio of 8:1:1. We report AUC-ROC and RMSE metrics for classification and regression tasks, respectively. The mean and standard deviations of results over three random seeds are provided by default.


\begin{table}[h]
  \resizebox{1.0 \columnwidth}{!}{
      \begin{tabular}{lccccccccc}
      \hline
      Methods                        & \multicolumn{6}{c}{Classification (AUC-ROC $\% \uparrow$ ) }                                     & \multicolumn{3}{c}{Regression (RMSE $\downarrow$)}             \\
      Dataset                        & BBBP      & BACE       & ClinTox   & Tox21     & ToxCast   & SIDER     & ESOL         & FreeSolv     & Lipo         \\
      \#Molecules                    & 2039      & 1513       & 1478      & 7831      & 8575      & 1427      & 1128         & 642          & 4200         \\
      \#Tasks                        & 1         & 1          & 2         & 12        & 617       & 27        & 1            & 1            & 1            \\ \hline
      D-MPNN                         & 71.0(0.3) & 80.9(0.6)  & 90.6(0.6) & 75.9(0.7) & 65.5(0.3) & 57.0(0.7) & 1.050(0.008) & 2.082(0.082) & 0.683(0.016) \\
      Attentive FP                   & 64.3(1.8) & 78.4(0.02) & 84.7(0.3) & 76.1(0.5) & 63.7(0.2) & 60.6(3.2) & 0.877(0.029) & 2.073(0.183) & 0.721(0.001) \\
      $\text{N-Gram}_{\text{RF}}$                     & 69.7(0.6) & 77.9(1.5)  & 77.5(4.0) & 74.3(0.4) & --        & 66.8(0.7) & 1.074(0.107) & 2.688(0.085) & 0.812(0.028) \\
      $\text{N-Gram}_{\text{XGB}}$                     & 69.1(0.8) & 79.1(1.3)  & 87.5(2.7) & 75.8(0.9) & --        & 65.5(0.7) & 1.083(0.082) & 5.061(0.744) & 2.072(0.030) \\
      MolCLR                         & 72.2(2.1) & 82.4(0.9)  & 91.2(3.5) & 75.0(0.2) & --        & 58.9(1.4) & 1.271(0.040) & 2.594(0.249) & 0.691(0.004) \\
      PretrainGNN                    & 68.7(1.3) & 84.5(0.7)  & 72.6(1.5) & 78.1(0.6) & 65.7(0.6) & 62.7(0.8) & 1.100(0.006) & 2.764(0.002) & 0.739(0.003) \\
      GraphMVP-G                     & 70.8(0.5) & 79.3(1.5)  & 79.1(2.8) & 75.9(0.5) & 63.1(0.2) & 60.2(1.1) & --           & --           & --           \\
      GraphMVP-C                     & 72.4(1.6) & 81.2(0.9)  & 76.3(1.9) & 74.4(0.2) & 63.1(0.4) & 63.9(1.2) & --           & --           & --           \\
      3DInfomax                      & 69.1(1.1) & 79.4(1.9)  & 59.4(3.2) & 74.5(0.7) & 64.4(0.9) & 53.4(3.3) & 0.894(0.028) & 2.34(0.227)  & 0.695(0.012) \\
      MICRO-graph                    & 77.2(2.0) & 84.4(1.1)  & 77.0(2.0) & 77.0(0.8) & 65.2(0.8) & 56.7(1.1) & --           & --           & --           \\
      $\text{GROVER}_{\text{base}}$  & 70.0(0.1) & 82.6(0.7)  & 81.2(3.0) & 74.3(0.1) & 65.4(0.4) & 64.8(0.6) & 0.983(0.090) & 2.176(0.052) & 0.817(0.008) \\
      $\text{GROVER}_{\text{large}}$ & 69.5(0.1) & 81.0(1.4)  & 76.2(3.7) & 73.5(0.1) & 65.3(0.5) & 65.4(0.1) & 0.895(0.017) & 2.272(0.051) & 0.823(0.010) \\
      GEM                            & 72.4(0.4) & 85.6(1.1)  & 90.1(1.3) & 78.1(0.1) & 69.2(0.4) & 67.2(0.4) & 0.798(0.029) & 1.877(0.094) & 0.660(0.008) \\
      Uni-Mol                        & 72.9(0.6) & 85.7(0.2)  & 91.9(1.8) & 79.6(0.5) & 69.6(0.1) & 65.9(1.3) & 0.788(0.029) & 1.620(0.035) & 0.603(0.010) \\ \hline
      CoSP                           & 73.1(0.3) & 84.3(1.1)  & 91.3(1.5) & 78.5(0.1) & 69.3(0.2) & 64.7(1.5) & 0.785(0.029) & 1.752(0.042) & 0.621(0.012) \\ \hline
      \end{tabular}}
      \caption{Molecule property prediction}
      \label{tab:property_pred}
\end{table}

\paragraph{Baselines} We evaluate CoSP against 14 baselines, including D-MPNN \citep{yang2019analyzing}, Attentive FP \citep{xiong2019pushing}, $\text{N-Gram}_{\text{RF}}$, $\text{N-Gram}_{\text{XGB}}$ \citep{liu2019n}, MolCLR \citep{wang2022molecular}, PretrainGNN \citep{hu2019strategies}, GraphMVP-G, GraphMVP-C \citep{liu2021pre}, 3DInfomax \citep{stark20213d}, MICRO-graph \citep{zhang2020motif}, $\text{GROVER}_{\text{base}}$, $\text{GROVER}_{\text{large}}$ \citep{rong2020self}, GEM \citep{fang2022geometry} and Uni-Mol \citep{zhou2022uni}. Except for $\text{N-Gram}_{\text{RF}}$ and $\text{N-Gram}_{\text{XGB}}$, all these baselines are pretraining methods. Some of the methods mentioned in the related works are not included because the experimental setup, e.g., data spliting, may be different.

\paragraph{Results and analysis} We show results in Table.\ref{tab:property_pred}. The main observations are:  (1) CoSP could not achieve the best results, probably because we used much less pre-training data than baseline methods. However, we have used complex data we could find, and more data will take a long time to accumulate. (2) CoSP achieves competitive results on both classification and regression and outperforms most of the baselines. These results indicate that the model could learn expressive molecular features for property prediction. With the increasing amount of complexes, we believes that the performance can be further improved.

\subsection{Downstream task 3: Virtual screening}
\paragraph{Experimental setup} Could the model distinguish molecules most likely to bind to the protein pockets? We evaluate CoSP on the  DUD-E \citep{mysinger2012directory} dataset which consists of 102 targets across different protein families. For each target, DUD-E provides 224 actives (positive examples) and 10,000 decoy ligands (negative examples) in average. The decoys were calculated by choosing them to be physically similar but topologically different from the actives. During finetuning, we use the same data splitting as GraphCNN \citep{torng2019graph}, and report the AUC-ROC and ROC enrichment (RE) scores. Note that $x\% \text{RE}$ indicates the ratio of the true positive rate (TPR) to the false positive rate (FPR) at $x\%$ FPR value.

\paragraph{Baselines} We compare CoSP with AutoDock Vina \citep{vina2010improving}, RF-score \citep{ballester2010machine}, NNScore \citep{durrant2010nnscore}, 3DCNN \citep{ragoza2017protein} and GraphCNN \citep{torng2019graph}. AutoDock Vina is an commonly used open-source program for doing molecular docking. RF-score use random forest capture protein-ligand binding effects. NNScore, 3DCNN, and GraphCNN use MLP, CNN and GNN to learn the protein-ligand binding, respectively.

\begin{table}[h]
  \small
  \centering
  \begin{tabular}{lccccc}
    \toprule
            & \multicolumn{5}{c}{DUD-E}                                                                                                                           \\
            & \multicolumn{1}{l}{AUC-ROC $\uparrow$} & \multicolumn{1}{l}{0.5\% RE $\uparrow$} & \multicolumn{1}{l}{1.0\% RE $\uparrow$} & \multicolumn{1}{l}{2.0\% RE $\uparrow$} & \multicolumn{1}{l}{5.0\% RE $\uparrow$} \\
  \midrule
  Vina      & 0.716                   & 9.139                        & 7.321                        & 5.881                        & 4.444                        \\
  RF-score  & 0.622                   & 5.628                        & 4.274                        & 3.499                        & 2.678                        \\
  NNScore   & 0.584                   & 4.166                        & 2.980                        & 2.460                        & 1.891                        \\
  Graph CNN & 0.886                   & 44.406                       & 29.748                       & 19.408                       & 10.735                       \\
  3DCNN     & 0.868                   & 42.559                       & 29.654                       & 19.363                       & 10.710                       \\ \hline
  CoSP      & 0.901   &  51.048        &   35.978       &  23.681      & 12.212   \\     
  \bottomrule
  \end{tabular}
  \caption{Virtual screening results on DUD-E}
  \label{tab:vs_results}
\end{table}

\paragraph{Results and analysis} We present results in Table.\ref{tab:vs_results}, and observe that: (1) Random forest and MLP-based RF-score and NNScore achieve competitive results to Vina, indicating the potential of machine learning in virtual screening. (2) Deeplearning-based Graph CNN and 3DCNN significantly outperforms both RF-score and NNScore. (3) CoSP provides the best results by achieving a 0.90 AUC score and outperforms baselines in RE scores.

\section{Conclusion}

This paper proposes a co-supervised pretraining framework to learn pocket and ligand spaces via contrastive learning jointly. The pretrained model could achieve competitive results on pocket matching, molecule property predictions, and virtual screening by injecting the interaction knowledge. We hope the unified modeling framework could further boost the development of AI drug discovery.

\bibliographystyle{plain}
\bibliography{references}

\end{document}